\documentclass[conference]{IEEEtran}
\IEEEoverridecommandlockouts
\usepackage{hyperref}
\usepackage{multirow}
\usepackage{amsfonts}
\usepackage[noadjust]{cite}
\usepackage{bm}
\usepackage{graphicx}
\usepackage{amsmath}

\usepackage[symbol]{footmisc}

\def\BibTeX{{\rm B\kern-.05em{\sc i\kern-.025em b}\kern-.08em
    T\kern-.1667em\lower.7ex\hbox{E}\kern-.125emX}}
\usepackage{algorithm}
\usepackage{algorithmic}

\title{On the Effectiveness of Incremental Training of Large Language Models}

\author{\IEEEauthorblockN{Miles Q. Li}
\IEEEauthorblockA{\textit{School of Information Studies} \\
\textit{McGill University}\\
Montreal, Canada \\
miles.qi.li@mail.mcgill.ca}
\and
\IEEEauthorblockN{Benjamin C. M. Fung*\thanks{* Corresponding author.}}
\IEEEauthorblockA{\textit{School of Information Studies} \\
\textit{McGill University}\\
Montreal, Canada \\
ben.fung@mcgill.ca}
\and
\IEEEauthorblockN{Shih-Chia Huang}
\IEEEauthorblockA{\textit{Department of Electronic Engineering} \\
\textit{National Taipei University of Technology}\\
Taipei, Taiwan \\
schuang@ntut.edu.tw}
}

\begin{document}

\maketitle

\begin{abstract}
Training large language models is a computationally intensive process that often requires substantial resources to achieve state-of-the-art results. Incremental layer-wise training has been proposed as a potential strategy to optimize the training process by progressively introducing layers, with the expectation that this approach would lead to faster convergence and more efficient use of computational resources. In this paper, we investigate the effectiveness of incremental training for LLMs, dividing the training process into multiple stages where layers are added progressively. Our experimental results indicate that while the incremental approach initially demonstrates some computational efficiency, it ultimately requires greater overall computational costs to reach comparable performance to traditional full-scale training. Although the incremental training process can eventually close the performance gap with the baseline, it does so only after significantly extended continual training. These findings suggest that incremental layer-wise training may not be a viable alternative for training large language models, highlighting its limitations and providing valuable insights into the inefficiencies of this approach.
\end{abstract}

\section{Introduction}

Training large language models (LLMs) has become a cornerstone of advancements in natural language processing (NLP), significantly impacted by improvements in model scaling and optimization techniques. Despite the success of models like GPTs~\cite{brown2020language} and BERT/RoBERTa~\cite{tenney2019bert,liu2019roberta}, scaling these models demands substantial resources, with training time and computational costs increasing significantly as the model size grows \cite{kaplan2020scaling, clark2022mixture,dubey2024llama}. Efficiently scaling LLMs is critical not only for reducing costs but also for making model training more accessible and environmentally sustainable. Incremental layer-wise training has been proposed as a method to potentially reduce these costs by progressively introducing layers~\cite{hinton2006fast}. This approach allows earlier parts of the model to stabilize while incrementally training additional layers, potentially leading to faster convergence and more efficient use of computational resources \cite{bengio2006greedy, fahlman1989cascade}. However, the effectiveness of this approach remains unclear, particularly concerning its ability to capture required long-range dependencies, as earlier studies indicate that such strategies may not fully generalize when trained incrementally \cite{borgeaud2021improving}.

The intuition behind incremental layer-wise training is rooted in the hierarchical learning process of large language models. In these models, lower layers often capture low-level linguistic features such as word embeddings, syntactic patterns, and local dependencies \cite{belinkov2017analyzing, tenney2019bert}. Higher layers, on the other hand, tend to model high-level abstractions like semantic relationships, contextual understanding, and long-range dependencies \cite{ voita2019bottom,geva2020transformer}. High-level features are essentially combinations of low-level ones, implying that effective learning of high-level representations relies on the prior learning of low-level features. Training all layers simultaneously might therefore be inefficient, as higher layers may struggle to learn meaningful patterns before the lower layers have stabilized their representations \cite{raghu2017svcca, zeiler2014visualizing}. By progressively adding layers, incremental training aims to mirror this natural progression, allowing each layer to specialize and stabilize before serving as the foundation for subsequent layers. This approach aligns with the way neural networks hierarchically construct representations, potentially leading to more effective learning and convergence \cite{raghu2017svcca}.

However, despite the intuitive appeal of incremental training, its effectiveness has not been thoroughly examined in the context of large-scale language models. While previous research has shown that gradually increasing the model's capacity or context size~\cite{dubey2024llama,yang2024qwen2} can be beneficial in some cases, the specific benefits of incrementally adding layers remain uncertain.

Our study addresses this gap by empirically evaluating incremental training in large-scale language models, analyzing computational efficiency, convergence behavior, and performance against traditional full-layer training. We compare the performance of models trained incrementally with those trained using a traditional approach, where all layers are optimized from the start.

Our findings indicate that, contrary to initial expectations, the incremental layer-wise training approach does not deliver significant benefits in terms of computational efficiency or performance. While incremental training can eventually reach comparable performance to traditional full-scale training, it does so only after a continual training period, resulting in a higher overall computational cost. Despite early-stage gains, these models require extensive fine-tuning to bridge the performance gap with the baseline, making the incremental approach a less practical choice for large language model training. These results underscore the limitations of incremental layer-wise training and provide insights into why it may not serve as an efficient alternative to traditional methods.

\section{Related Work}

The pursuit of efficient training methods for large-scale neural networks has been an active area of research. Incremental or layer-wise training strategies have been explored in various contexts, aiming to reduce computational costs and memory requirements.

\subsection{Incremental Training in Deep Learning}

Incremental training, also known as layer-wise training or progressive stacking, has been applied in deep learning to gradually build up network architectures. Early work by Hinton et al. \cite{hinton2006fast} introduced a fast learning algorithm for deep belief nets, where layers are trained sequentially in an unsupervised manner while keeping the weights of previous layers fixed. Similarly, Bengio et al. \cite{bengio2006greedy} proposed Greedy Layer-Wise Training for deep networks, demonstrating that such approaches can initialize deep networks effectively.

Moreover, the Cascade-Correlation learning architecture \cite{fahlman1989cascade} incrementally builds neural networks by adding hidden units one at a time, freezing the weights of previously added units. This method aimed to overcome challenges in training deeper networks by simplifying the optimization problem.

While these approaches showed promise in certain settings, particularly in unsupervised pre-training and for shallower networks, they often struggled to match the performance of end-to-end training in supervised tasks for deeper architectures like modern LLMs. The inability to fully capture complex hierarchical representations when layers are trained incrementally has been a consistent challenge \cite{erhan2010does}.

\subsection{Efficient Training Techniques for LLMs}

Various methods have been proposed to improve the efficiency of training LLMs:
\begin{itemize}
\item Model Pruning: Reducing the number of parameters by removing redundant weights \cite{frankle2018lottery}.
\item Knowledge Distillation: Training smaller models to replicate the performance of larger ones \cite{hinton2015distilling}.
\item Mixed-Precision Training: Utilizing lower numerical precision to speed up computations \cite{micikevicius2017mixed}.
\item Layer Freezing: Training only a subset of layers while keeping others fixed \cite{howard2018universal}.
\end{itemize}

However, these methods come with trade-offs between efficiency and model performance, and their applicability to incremental training remains limited.

\subsection{Progressive Neural Networks}

Progressive neural networks \cite{rusu2016progressive} introduce new columns (networks) when learning new tasks, while keeping previous columns fixed to retain prior knowledge. This approach is beneficial in transfer learning and continual learning scenarios but differs from the incremental layer-wise training of a single task.

\subsection{Cognitive and Biological Inspirations}

Incremental learning is reminiscent of how humans and animals learn, gradually building upon prior knowledge. However, replicating this process in artificial neural networks has proven challenging due to issues like catastrophic forgetting and optimization difficulties \cite{goodfellow2013empirical}.

\section{Methodology}

Building upon earlier approaches that incrementally construct neural networks \cite{fahlman1989cascade, bengio2006greedy}, our goal is to assess whether progressively adding layers during training can improve computational efficiency and model performance in the context of modern LLMs.

This incremental approach is motivated by the understanding that higher-level layers depend on the representations learned by lower-level layers. Since high-level features are combinations of low-level ones, training higher layers before the lower layers have adequately learned foundational features may be ineffective and could lead to wasted computational resources \cite{yosinski2014transferable}. By first training the lower layers to capture basic linguistic features, we provide a stable and informative input for the higher layers to build upon. This approach also addresses the issue of internal covariate shift, as lower layers have sufficient time to stabilize their representations before training progresses to higher layers. This stabilization can reduce the shifting of input distributions for higher layers, leading to more effective optimization~\cite{ioffe2015batch}. Training the newly added layers in isolation allows them to adapt to the established representations from earlier layers without the interference of simultaneous updates throughout the entire network. This sequential learning process aims to optimize computational resources by avoiding unnecessary computations in higher layers during the early stages of training. The subsequent fine-tuning phase then harmonizes the representations across all trained layers, integrating the newly learned features with the existing model structure. Unlike previous methods that focused on unsupervised pre-training or shallow networks \cite{hinton2006fast, bengio2006greedy}, we aim to investigate whether this method provides any practical advantages for deep, transformer-based architectures by examining convergence speed, memory usage, and generalization ability.

\subsection{Model Architecture and Notation} 

Let the total number of layers in the LLM be denoted by $ L $, where $ L = n \times m $, with $ s $ being the total number of stages and $ m $ the number of layers added in each stage. We denote the layers at stage $ i $ as $ \mathcal{L}_i = \{l_{(i-1) \times m + 1}, l_{(i-1) \times m + 2}, \dots, l_{i \times m} \} $. For both incremental training and baseline training, we use the same architecture, dataset, and hyperparameters to ensure a fair comparison.

\subsection{Stage-wise Training Process}

Each stage $ i $ consists of the following two phases, which are designed to evaluate the potential benefits of training newly added layers in isolation before fine-tuning the entire model.

\subsubsection{Phase 1: Training New Layers}

During this phase, only the newly added layers $ \mathcal{L}_i $ are trained while keeping the parameters of all preceding layers $ \{ \mathcal{L}_1, \mathcal{L}_2, \dots, \mathcal{L}_{i-1} \} $ fixed. The motivation behind this approach is to isolate the training of new layers and prevent the random initialization from negatively impacting the performance of the previously trained parameters. However, our findings suggest that this isolation may not allow the model to sufficiently integrate newly learned features across different layers, which could hinder generalization. The optimization problem for this phase can be formulated as:

\[
\min_{\theta_i} \mathcal{L}(\theta_i; \theta_{1:(i-1)}, \mathcal{D})
\]

where $ \theta_i $ represents the parameters of the newly added layers $ \mathcal{L}_i $, $ \theta_{1:(i-1)} $ denotes the fixed parameters of the previously trained layers, and $ \mathcal{D} $ is the training data.

\begin{figure*}[htbp]
    \centering
    \includegraphics[width=18cm]{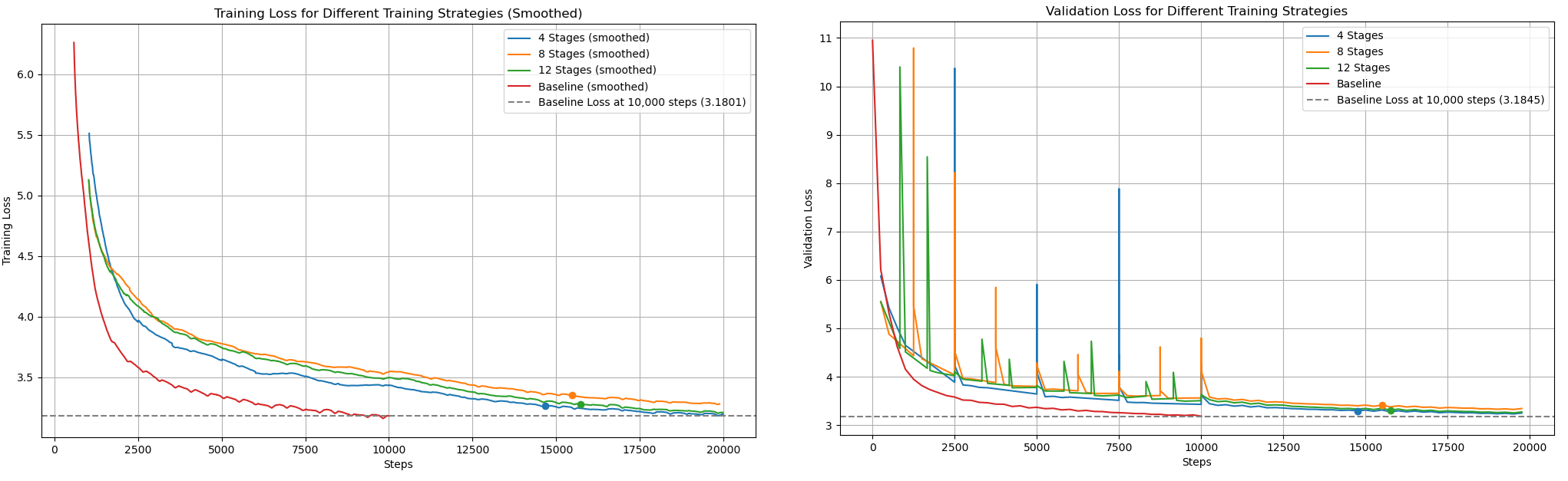}
    \caption{Training and validation loss curves comparing incremental layer-wise training (with 4, 8, and 12 stages) and baseline training. The large solid circles mark the points where the incremental training regimes have reached the same cumulative computational cost as the baseline model trained for 10,000 steps.}
    \label{fig:training_validation_loss}
\end{figure*}

\subsubsection{Phase 2: Fine-tuning All Layers}

After training the new layers, the entire model, consisting of layers $ \{ \mathcal{L}_1, \mathcal{L}_2, \dots, \mathcal{L}_i \} $, is fine-tuned together. The goal of this phase is to integrate the newly learned features into the existing model and improve overall optimization. Although this step aims to harmonize the learned features across all layers, our experimental results show that it may not be sufficient to close the performance gap with models trained using a traditional approach. The optimization problem during this phase is given by:

\[
\min_{\theta_{1:i}} \mathcal{L}(\theta_{1:i}; \mathcal{D})
\]

where $ \theta_{1:i} $ represents the parameters of all layers up to stage $ i $. 

\subsection{Incremental Layer Addition}

This process of introducing new layers and fine-tuning continues until all $ L $ layers have been trained. In our experiments, we varied the number of stages (e.g., 4, 8, and 12 stages) to examine whether the granularity of layer addition impacts the final performance. Optinally, context size and batch size can also be increased throughout the stages as the training is plateaued~\cite{dubey2024llama,yang2024qwen2}, following common practices aimed at enhancing model capabilities for longer dependencies.

\subsection{Continual Training Phase}
After completing the incremental training phases, an optional continual training phase can be applied to further improve the model's performance. In this phase, all model layers are optimized jointly, similar to traditional full-layer training. This continual training aims to integrate the learned representations across all layers, potentially closing any performance gaps observed during incremental training. The continual phase is intended to harmonize layer interactions and enhance generalization capabilities beyond what is achievable through isolated layer-wise training.

\subsection{Traditional Training Regime For Comparison}

The effectiveness of the incremental strategy should be compared against a the traditional full-layer training approach where all layers are trained simultaneously from the beginning. Both the baseline and incremental models share the same hyperparameters, architecture, and dataset. The primary difference is that the baseline approach trains all $ L $ layers together for the entire duration, while the incremental approach progressively adds layers. This comparison allows us to quantify the trade-offs between computational efficiency and model performance.

\subsection{Computational Cost Analysis}

In this subsection, we analyze the computational cost of incremental layer-wise training compared to traditional full-layer training. Our goal is to determine how many additional tokens of continual training are needed, as a ratio of the baseline training tokens $T$, so that the total computational cost of incremental training plus continual training equals the computational cost of the baseline training on $T$ tokens.

\subsubsection{Definitions and Assumptions}

Let:

\begin{itemize}
    \item $L$ be the total number of layers in the model.
    \item $S$ be the total number of stages in incremental training.
    \item $m$ be the number of layers added per stage, calculated as $m = \frac{L}{S}$ (assuming $L$ is divisible by $S$).
    \item $L_i$ be the total number of layers up to stage $i$, so $L_i = i \times m$.
    \item $T$ be the total number of tokens used in baseline training.
    \item $T_{\text{inc}}$ be the number of tokens used during the incremental training stages.
    \item $T_{\text{cont}}$ be the number of tokens used during the continual training phase.
    \item $c$ be the computational cost per layer per token for the forward or backward pass.
\end{itemize}

\textbf{Phases of Incremental Training}:

Each stage consists of two phases:

\begin{itemize}
    \item \textbf{Phase 1}: Train the newly added layers for $\frac{T_{\text{inc}}}{2S}$ tokens.
    \begin{itemize}
        \item \textbf{Forward pass}: Involves all layers up to the current stage ($L_i$ layers).
        \item \textbf{Backward pass}: Involves only the newly added layers ($m$ layers).
    \end{itemize}
    \item \textbf{Phase 2}: Fine-tune all layers up to the current stage for $\frac{T_{\text{inc}}}{2S}$ tokens.
    \begin{itemize}
        \item \textbf{Forward pass}: Involves all layers up to the current stage ($L_i$ layers).
        \item \textbf{Backward pass}: Involves all layers up to the current stage ($L_i$ layers).
    \end{itemize}
\end{itemize}

\textbf{Assumptions}:

\begin{itemize}
    \item The computational cost per token is directly proportional to the number of layers involved in the forward and backward passes.
    \item The cost per layer per token is the same for both the forward and backward passes.
    \item The computational cost during the continual training phase is the same per token as in the baseline training, since all layers are involved in both forward and backward passes.
\end{itemize}

\subsubsection{Computational Cost of Baseline Training}

For the baseline model:

\[
\text{Cost per token} = L \times c_{\text{forward}} + L \times c_{\text{backward}} = 2L \times c
\]

Total computational cost:

\[
C_{\text{baseline}} = T \times 2L \times c = 2TLc
\]

\subsubsection{Computational Cost of Incremental Training}

The total computational cost of incremental training includes the costs from all stages and phases.

\textbf{Phase 1 of Stage $i$}:

\begin{itemize}
    \item Tokens processed: $\frac{T_{\text{inc}}}{2S}$
    \item Forward pass cost per token: $L_i \times c$
    \item Backward pass cost per token: $m \times c$
    \item Total cost per token: $(L_i + m) \times c$
    \item Total cost: $C_{\text{phase1}_i} = \frac{T_{\text{inc}}}{2S} \times (L_i + m) \times c$
\end{itemize}

\textbf{Phase 2 of Stage $i$}:

\begin{itemize}
    \item Tokens processed: $\frac{T_{\text{inc}}}{2S}$
    \item Forward pass cost per token: $L_i \times c$
    \item Backward pass cost per token: $L_i \times c$
    \item Total cost per token: $2L_i \times c$
    \item Total cost: $C_{\text{phase2}_i} = \frac{T_{\text{inc}}}{2S} \times 2L_i \times c$
\end{itemize}

\textbf{Total Cost for Stage $i$}:

\[
C_{\text{stage}_i} = C_{\text{phase1}_i} + C_{\text{phase2}_i} = \frac{T_{\text{inc}}}{2S} \times (3L_i + m) \times c
\]

\textbf{Total Incremental Training Cost}:

\[
C_{\text{incremental}} = \sum_{i=1}^{S} C_{\text{stage}_i} = \frac{T_{\text{inc}} c}{2S} \sum_{i=1}^{S} (3L_i + m)
\]

Since $L_i = i m$, we have:

\[
\sum_{i=1}^{S} (3L_i + m) = 3m \sum_{i=1}^{S} i + m S = m S \left( \frac{3S + 5}{2} \right)
\]

Therefore, the total incremental training cost is:

\[
C_{\text{incremental}} = \frac{T_{\text{inc}} c}{2S} \times m S \times \left( \frac{3S + 5}{2} \right) = \frac{T_{\text{inc}} c m}{2} \times \left( \frac{3S + 5}{2} \right)
\]

Since $m = \frac{L}{S}$:

\[
C_{\text{incremental}} = \frac{T_{\text{inc}} c L}{2S} \times \left( \frac{3S + 5}{2} \right)
\]

Simplify:

\[
C_{\text{incremental}} = \frac{T_{\text{inc}} c L (3S + 5)}{4S}
\]

\subsubsection{Computational Cost of Continual Training}

The computational cost per token during continual training is the same as the baseline:

\[
\text{Cost per token} = 2L \times c
\]

Total computational cost:

\[
C_{\text{continual}} = T_{\text{cont}} \times 2L \times c = 2T_{\text{cont}} L c
\]

\subsubsection{Total Computational Cost and Equality with Baseline}

The total computational cost of the incremental approach is:

\[
C_{\text{total}} = C_{\text{incremental}} + C_{\text{continual}}
\]

We set $C_{\text{total}} = C_{\text{baseline}}$ to find $T_{\text{cont}}$:

\[
C_{\text{incremental}} + C_{\text{continual}} = C_{\text{baseline}}
\]
\[
\frac{T_{\text{inc}} c L (3S + 5)}{4S} + 2T_{\text{cont}} L c = 2T L c
\]

Since $T_{\text{inc}} = T$:

\[
\frac{T c L (3S + 5)}{4S} + 2T_{\text{cont}} L c = 2T L c
\]

Solve for $T_{cont}$

\[
T_{\text{cont}} = \frac{5}{8}(1-\frac{1}{S})T
\]

This formula allows us to calculate the required amount of continual training (as a ratio of $T$) to equal the computational cost of the baseline training.

\section{Experiments}

The goal of our experiments is to empirically evaluate the effectiveness of incremental layer-wise training for large language models and compare it against traditional full-scale training. We focus on examining whether incremental training offers any advantages in terms of computational efficiency or model performance. Our results indicate that the incremental approach struggles to match the baseline performance, even when utilizing similar or greater computational resources. 

\subsection{Experimental Setup}
To ensure a fair comparison, we used the GPT-2 architecture with 124.4 million parameters for both the baseline and incremental training regimes. The training data consisted of 10 billion tokens from the FineWeb-edu dataset~\cite{lozhkov2024fineweb-edu}. The models were trained using the AdamW optimizer~\cite{loshchilov2019adamw} with a learning rate of 6e-4, weight decay of 0.1, using a batch size of 512 sequences, each with a sequence length of 1,024 tokens, totaling 524,288 tokens per batch. 

For the incremental training experiments, the total number of layers of GPT-2 is 12. They were divided into various configurations of stages, such as 4, 8, and 12 stages. Each stage involved two phases: training the newly added layers (Phase 1) while keeping the previous layers fixed, followed by fine-tuning all layers up to the current stage (Phase 2). 

The baseline model was trained by optimizing all layers simultaneously from the start using the same architecture and hyperparameters. This approach allows us to directly compare the outcomes of the incremental training against traditional full-scale training.

In our experiments, we applied incremental training for the first 10,000 steps, after which we transitioned to continual training where all parameters were optimized jointly as in the traditional training process. We selected training with the baseline method in 10,000 steps as the performance benchmark, allowing us to assess the incremental model's ability to match baseline performance after this continual training phase. The incremnental training regime with continual training reaches the same computational budget according to our formula in the previous section at 14,688 steps for 4 stages, 15,469 steps for 8 stages, and 15,729 steps for 12 stages. This setup enabled a fair comparison between the two approaches, focusing on whether the incremental model could achieve similar generalization within a comparable training duration.

\subsection{Evaluation Metrics}
To evaluate the models, we monitored three primary metrics: training loss, validation loss, and accuracy on the HellaSwag benchmark~\cite{zellers2019hellaswag}. The training and validation loss were used to assess convergence behavior, while the HellaSwag benchmark provided insights into the generalization capabilities of the models.

\begin{figure}[h]
    \centering
    \includegraphics[width=9.5cm]{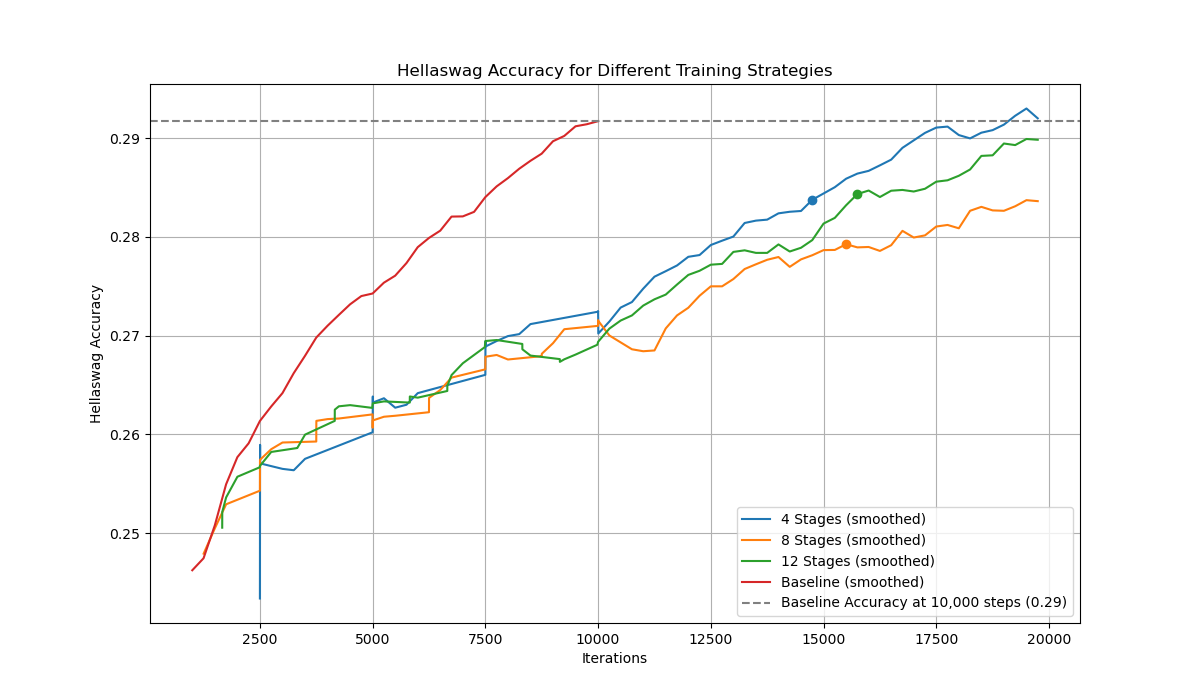} 
    \caption{HellaSwag accuracy scores comparing incremental layer-wise training (with 4, 8, and 12 stages) and baseline training. The large solid circles indicate the performance of the incremental models at the steps where their cumulative computational cost equals that of the baseline model trained for 10,000 steps.}
    \label{fig:hellaswag_accuracy}
\end{figure}

\subsection{Results}

\subsubsection{Training and Validation Loss}

Figure~\ref{fig:hellaswag_accuracy} presents the training and validation loss curves for both the baseline and incremental models. The baseline model exhibits faster convergence with lower overall training and validation losses throughout the training process. In contrast, the incremental models show higher losses, indicating slower convergence and suboptimal performance at equivalent computational budgets.

At the points where the incremental models have expended the same cumulative computational cost as the baseline model trained for 10,000 steps (marked by the large solid circles in the figure), all incremental models display higher training and validation losses compared to the baseline. Specifically, the four-stage incremental model still lags behind the baseline in terms of loss values at this computational budget. Although the four-stage incremental model eventually reaches training and validation losses comparable to the baseline, it does so only after significantly more training steps, highlighting that the incremental approach requires substantially more computational effort to achieve similar performance.

\subsubsection{HellaSwag Benchmark Evaluation}

Figure~\ref{fig:training_validation_loss} presents the accuracy scores on the HellaSwag benchmark for both the baseline and incremental training regimes. The baseline model consistently outperforms the incremental training regimes throughout the training process. At the points of equal cumulative computational cost (indicated by the large solid circles), the incremental models show significantly lower accuracy compared to the baseline trained for 10,000 steps. The four-stage incremental model, for instance, demonstrates a notable performance gap at this point.

While the four-stage incremental model eventually closes the accuracy gap with the baseline, it requires approximately  much more than the baseline's computational budget—to achieve comparable performance. This extended training underscores that the incremental approach demands substantially more computational resources to match the baseline's generalization capabilities on the HellaSwag benchmark.

\subsection{Analysis of Results}

The experimental results indicate that the incremental layer-wise training approach underperforms compared to the baseline when evaluated at the same computational budget. Despite the initial reduction in computational cost per step during the early stages of incremental training, the models require a significant amount of additional continual training to match the total computational cost of the baseline model.

At the points where the cumulative computational costs are equal, all incremental models exhibit higher training and validation losses and lower HellaSwag accuracy than the baseline. The four-stage incremental model eventually achieves performance similar to the baseline, but only after substantially more training steps and computational resources. This prolonged process suggests that incremental training does not offer practical benefits over traditional training, as the additional resources required outweigh the early efficiency gains.

\section{Discussion}

In this section, we discuss the implications of our findings and discuss assumptions made in our computational cost analysis.

\subsection{Computational Efficiency}

While incremental training reduces memory usage and computational cost per step in the early stages, achieving performance comparable to the baseline ultimately requires significantly more continual training. The initial computational savings are offset by the extended training time and additional resources needed during the continual training phase. At the same cumulative computational cost, incremental models perform worse than the baseline, indicating lower computational efficiency. This makes the incremental approach less practical for large language model training, as it necessitates additional resources to achieve results similar to traditional full-layer training.

\subsection{Incremental Strategies Not Explored}

While incrementing batch size and context length as we increase the number of layers could be a potential direction to explore, we did not pursue this approach. Our findings indicated that even without reducing the number of tokens in the batch during the early stages, the incremental training results were not satisfactory. So, there is no point in exploring on that direction further in this study.

\subsection{Assumption on Compute Cost of Forward and Backward Passes}

In our computational cost analysis, we assumed that the computational cost per layer per token is the same for both the forward and backward passes. We acknowledge that this assumption is not entirely accurate, as the backward pass generally requires more computational resources due to gradient computations and the storage of intermediate activations for backpropagation~\cite{goodfellow2016deep}. However, this simplification does not significantly affect our overall conclusions. Even when accounting for the higher computational cost of the backward pass, the incremental training regime still demands substantial continual training to approximate the performance of the traditional baseline. This continual training phase consumes computational resources that are close to the compute budget of the traditional training approach. Therefore, despite the inaccuracy in the initial assumption, our fundamental finding remains valid: incremental layer-wise training does not offer computational efficiency advantages over full-layer training for large language models.

\section{Conclusion}

In this paper, we evaluated the effectiveness of incremental layer-wise training for large language models. Our findings demonstrate that, contrary to expectations, this approach does not offer benefits in computational efficiency or model performance. Incremental training regimes underperform traditional full-layer training, even when accounting for the same cumulative computational cost. The need for extensive continual training to match baseline performance makes the incremental approach less practical for large language model training. These results highlight the limitations of incremental layer-wise training and underscore the importance of exploring alternative methods for efficient LLM training.

\bibliography{sampl}

\begin{thebibliography}{10}
\providecommand{\url}[1]{#1}
\csname url@samestyle\endcsname
\providecommand{\newblock}{\relax}
\providecommand{\bibinfo}[2]{#2}
\providecommand{\BIBentrySTDinterwordspacing}{\spaceskip=0pt\relax}
\providecommand{\BIBentryALTinterwordstretchfactor}{4}
\providecommand{\BIBentryALTinterwordspacing}{\spaceskip=\fontdimen2\font plus
\BIBentryALTinterwordstretchfactor\fontdimen3\font minus
  \fontdimen4\font\relax}
\providecommand{\BIBforeignlanguage}[2]{{%
\expandafter\ifx\csname l@#1\endcsname\relax
\typeout{** WARNING: IEEEtran.bst: No hyphenation pattern has been}%
\typeout{** loaded for the language `#1'. Using the pattern for}%
\typeout{** the default language instead.}%
\else
\language=\csname l@#1\endcsname
\fi
#2}}
\providecommand{\BIBdecl}{\relax}
\BIBdecl

\bibitem{brown2020language}
T.~B. Brown, ``Language models are few-shot learners,'' \emph{arXiv preprint
  arXiv:2005.14165}, 2020.

\bibitem{tenney2019bert}
I.~Tenney, ``Bert rediscovers the classical nlp pipeline,'' \emph{arXiv
  preprint arXiv:1905.05950}, 2019.

\bibitem{liu2019roberta}
Y.~Liu, ``Roberta: A robustly optimized bert pretraining approach,''
  \emph{arXiv preprint arXiv:1907.11692}, vol. 364, 2019.

\bibitem{kaplan2020scaling}
J.~Kaplan, S.~McCandlish, T.~Henighan, T.~Brown, B.~Chess, R.~Child, S.~Gray,
  A.~Radford, J.~Wu, and D.~Amodei, ``Scaling laws for neural language
  models,'' \emph{arXiv preprint arXiv:2001.08361}, 2020.

\bibitem{clark2022mixture}
C.~Clark, M.~Jovanovic, and P.~Voss, ``Training compute-optimal large language
  models,'' \emph{arXiv preprint arXiv:2203.15556}, 2022.

\bibitem{dubey2024llama}
A.~Dubey, A.~Jauhri, A.~Pandey, A.~Kadian, A.~Al-Dahle, A.~Letman, A.~Mathur,
  A.~Schelten, A.~Yang, A.~Fan \emph{et~al.}, ``The llama 3 herd of models,''
  \emph{arXiv preprint arXiv:2407.21783}, 2024.

\bibitem{hinton2006fast}
G.~E. Hinton, S.~Osindero, and Y.-W. Teh, ``A fast learning algorithm for deep
  belief nets,'' \emph{Neural computation}, vol.~18, no.~7, pp. 1527--1554,
  2006.

\bibitem{bengio2006greedy}
Y.~Bengio, P.~Lamblin, D.~Popovici, and H.~Larochelle, ``Greedy layer-wise
  training of deep networks,'' \emph{Advances in neural information processing
  systems}, vol.~19, 2006.

\bibitem{fahlman1989cascade}
S.~Fahlman and C.~Lebiere, ``The cascade-correlation learning architecture,''
  \emph{Advances in neural information processing systems}, vol.~2, 1989.

\bibitem{borgeaud2021improving}
S.~Borgeaud, A.~Mensch, J.~Hoffmann, T.~Cai, E.~Rutherford, K.~Millican,
  G.~van~den Driessche, J.-B. Lespiau, B.~Damoc, A.~Clark \emph{et~al.},
  ``Improving language models by retrieving from trillions of tokens,''
  \emph{arXiv preprint arXiv:2112.04426}, 2021.

\bibitem{belinkov2017analyzing}
Y.~Belinkov and J.~Glass, ``Analyzing hidden representations in end-to-end
  automatic speech recognition systems,'' \emph{Advances in Neural Information
  Processing Systems}, vol.~30, 2017.

\bibitem{voita2019bottom}
E.~Voita, R.~Sennrich, and I.~Titov, ``The bottom-up evolution of
  representations in the transformer: A study with machine translation and
  language modeling objectives,'' \emph{arXiv preprint arXiv:1909.01380}, 2019.

\bibitem{geva2020transformer}
M.~Geva, R.~Schuster, J.~Berant, and O.~Levy, ``Transformer feed-forward layers
  are key-value memories,'' \emph{arXiv preprint arXiv:2012.14913}, 2020.

\bibitem{raghu2017svcca}
M.~Raghu, J.~Gilmer, J.~Yosinski, and J.~Sohl-Dickstein, ``Svcca: Singular
  vector canonical correlation analysis for deep learning dynamics and
  interpretability,'' \emph{Advances in neural information processing systems},
  vol.~30, 2017.

\bibitem{zeiler2014visualizing}
M.~Zeiler, ``Visualizing and understanding convolutional networks,'' in
  \emph{European conference on computer vision/arXiv}, vol. 1311, 2014.

\bibitem{yang2024qwen2}
A.~Yang, B.~Yang, B.~Hui, B.~Zheng, B.~Yu, C.~Zhou, C.~Li, C.~Li, D.~Liu,
  F.~Huang \emph{et~al.}, ``Qwen2 technical report,'' \emph{arXiv preprint
  arXiv:2407.10671}, 2024.

\bibitem{erhan2010does}
D.~Erhan, A.~Courville, Y.~Bengio, and P.~Vincent, ``Why does unsupervised
  pre-training help deep learning?'' in \emph{Proceedings of the thirteenth
  international conference on artificial intelligence and statistics}.\hskip
  1em plus 0.5em minus 0.4em\relax JMLR Workshop and Conference Proceedings,
  2010, pp. 201--208.

\bibitem{frankle2018lottery}
J.~Frankle and M.~Carbin, ``The lottery ticket hypothesis: Finding sparse,
  trainable neural networks,'' \emph{arXiv preprint arXiv:1803.03635}, 2018.

\bibitem{hinton2015distilling}
G.~Hinton, ``Distilling the knowledge in a neural network,'' \emph{arXiv
  preprint arXiv:1503.02531}, 2015.

\bibitem{micikevicius2017mixed}
P.~Micikevicius, S.~Narang, J.~Alben, G.~Diamos, E.~Elsen, D.~Garcia,
  B.~Ginsburg, M.~Houston, O.~Kuchaiev, G.~Venkatesh \emph{et~al.}, ``Mixed
  precision training,'' \emph{arXiv preprint arXiv:1710.03740}, 2017.

\bibitem{howard2018universal}
J.~Howard and S.~Ruder, ``Universal language model fine-tuning for text
  classification,'' \emph{arXiv preprint arXiv:1801.06146}, 2018.

\bibitem{rusu2016progressive}
A.~A. Rusu, N.~C. Rabinowitz, G.~Desjardins, H.~Soyer, J.~Kirkpatrick,
  K.~Kavukcuoglu, R.~Pascanu, and R.~Hadsell, ``Progressive neural networks,''
  \emph{arXiv preprint arXiv:1606.04671}, 2016.

\bibitem{goodfellow2013empirical}
I.~J. Goodfellow, M.~Mirza, D.~Xiao, A.~Courville, and Y.~Bengio, ``An
  empirical investigation of catastrophic forgetting in gradient-based neural
  networks,'' \emph{arXiv preprint arXiv:1312.6211}, 2013.

\bibitem{yosinski2014transferable}
J.~Yosinski, J.~Clune, Y.~Bengio, and H.~Lipson, ``How transferable are
  features in deep neural networks?'' \emph{Advances in neural information
  processing systems}, vol.~27, 2014.

\bibitem{ioffe2015batch}
S.~Ioffe, ``Batch normalization: Accelerating deep network training by reducing
  internal covariate shift,'' \emph{arXiv preprint arXiv:1502.03167}, 2015.

\bibitem{lozhkov2024fineweb-edu}
\BIBentryALTinterwordspacing
A.~Lozhkov, L.~Ben~Allal, L.~von Werra, and T.~Wolf, ``Fineweb-edu,'' May 2024.
  [Online]. Available:
  \url{https://huggingface.co/datasets/HuggingFaceFW/fineweb-edu}
\BIBentrySTDinterwordspacing

\bibitem{loshchilov2019adamw}
I.~Loshchilov and F.~Hutter, ``Decoupled weight decay regularization,''
  \emph{arXiv preprint arXiv:1711.05101}, 2019.

\bibitem{zellers2019hellaswag}
R.~Zellers, A.~Holtzman, Y.~Bisk, A.~Farhadi, and Y.~Choi, ``Hellaswag: Can a
  machine really finish your sentence?'' \emph{arXiv preprint
  arXiv:1905.07830}, 2019.

\bibitem{goodfellow2016deep}
I.~Goodfellow, Y.~Bengio, and A.~Courville, ``Deep feedforward networks,''
  \emph{Deep learning}, no.~1, 2016.

\end{thebibliography}

\bibliographystyle{IEEEtran}
\vspace{12pt}

\end{document}